\documentclass[10pt,twocolumn,letterpaper]{article}
\usepackage{cvpr}
\usepackage{times}
\usepackage{epsfig}
\usepackage{graphicx}
\usepackage{amsmath}
\usepackage{amssymb}
\usepackage{bm}
\usepackage{algorithm}
\usepackage{algorithmic}
\usepackage{comment}
\usepackage{authblk}
\usepackage{url}
\usepackage{array}

\usepackage[breaklinks=true,bookmarks=false]{hyperref}

\cvprfinalcopy 

\def\cvprPaperID{2464} 

\ifcvprfinal\fi
\begin{document}

\title{Alive Caricature from 2D to 3D}


\author{Qianyi Wu$^{1}$, Juyong Zhang$^{1}$\thanks{Corresponding author: Juyong Zhang}, Yu-Kun Lai$^{2}$, Jianmin Zheng$^{3}$ and   Jianfei Cai$^{3}$ \\
        $^{1}$University of Science and Technology of China, China\\
        $^{2}$Cardiff University, UK\\
        $^{3}$Nanyang Technological University, Singapore\\
        \tt\small{wqy9619@mail.ustc.edu.cn, 		 juyong@ustc.edu.cn,
        Yukun.Lai@cs.cardiff.ac.uk, 
         \{ASJMZheng,ASJFCai\}@ntu.edu.sg}
}




\maketitle

\begin{abstract}
Caricature is an art form that expresses subjects in abstract, simple and exaggerated views. While many caricatures are 2D images, this paper presents an algorithm for creating expressive 3D caricatures from 2D caricature images with minimum user interaction. The key idea of our approach is to introduce an intrinsic deformation representation that has the capability of extrapolation, enabling us to create a deformation space from standard face datasets, which maintains face constraints and meanwhile is sufficiently large for producing exaggerated face models. Built upon the proposed deformation representation, an optimization model is formulated to find the 3D caricature that captures the style of the 2D caricature image automatically. The experiments show that our approach has better capability in expressing caricatures than those fitting approaches directly using classical parametric face models such as 3DMM and FaceWareHouse. Moreover, our approach is based on standard face datasets and avoids constructing complicated 3D caricature training sets, which provides great flexibility in real applications.

\end{abstract}

\section{Introduction}

\begin{figure}[tp]
\begin{center}
   \includegraphics[width=1\linewidth]{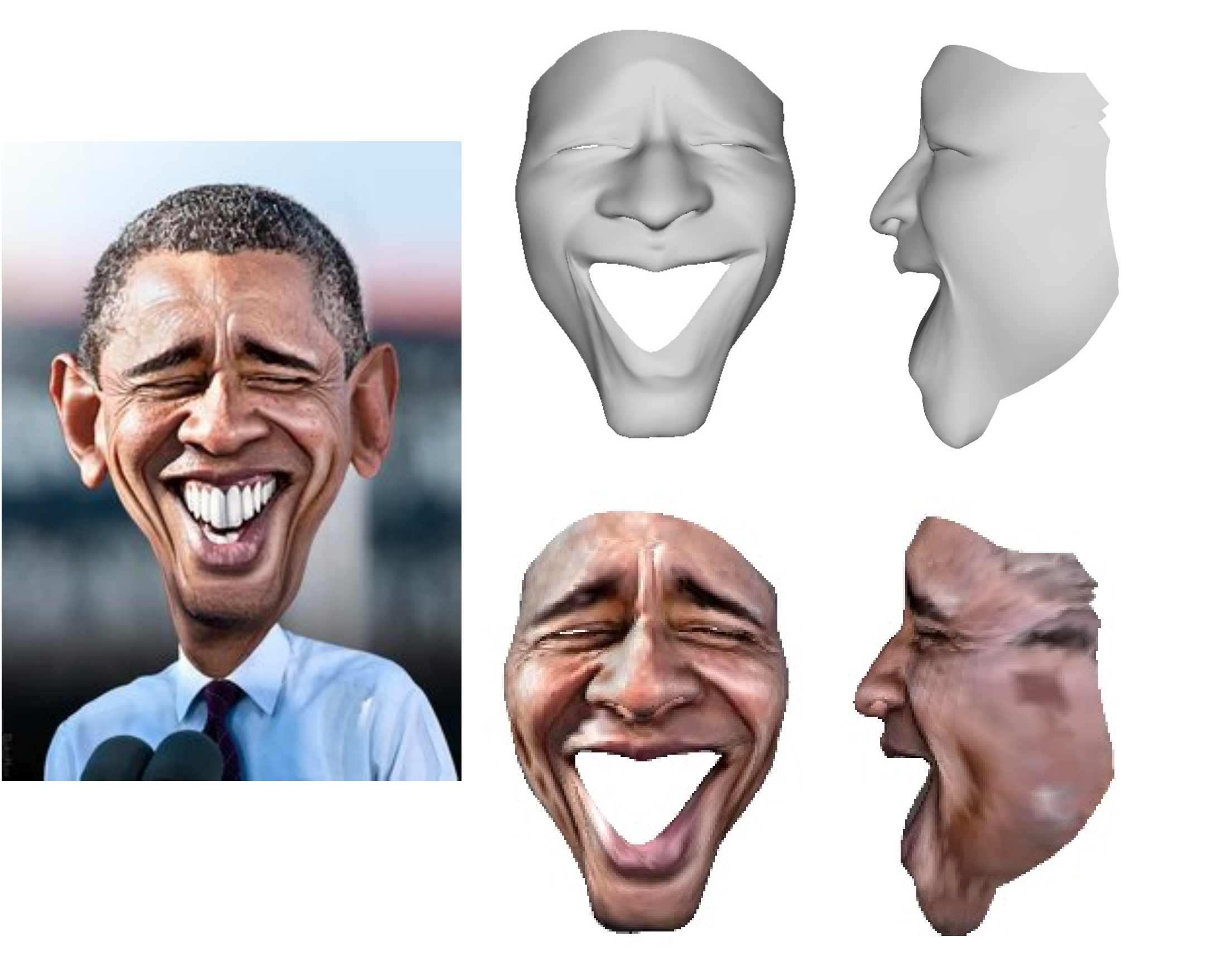}
\end{center}
\caption{An example of 3D caricature generation from a single image. Given a single caricature image (left), our algorithm generates its 3D model and the model with texture (right, displayed in two views).}
\label{fig1}
\end{figure}
Caricature is a pictorial representation or description that deliberately exaggerates a person's distinctive features or peculiarities to create an easily identifiable visual likeness with a comic effect~\cite{Sadimon10survey}. This vivid art form contains the concepts of abstraction, simplification and exaggeration. It has been shown that the effect of producing caricature can increase face recognition rates~\cite{RHODES1987473,CVPR91,cvpr99}. Since Brennan presented the first interactive caricature generator in 1985~\cite{Brennan85}, many approaches and computer-assisted caricature generation systems have been developed~\cite{liang2002example,Shum-CVPR03,liao2004automatic,CGF:CGF1804}. Most of these works focus on 2D caricature generation. Our goal is to develop techniques for creating 3D caricatures from 2D caricature images. Such expressive 3D models of caricatures are interesting and useful, for example, in cartoon and social media.

Creating 3D caricatures from 2D images is a problem of image-based modeling. A closely-related and very interesting problem is face reconstruction which is widely studied in computer vision. Due to diverse geometric and texture variations and a large variety of identities and expressions, face reconstruction is a nontrivial task. Recently face reconstruction has achieved great progress. Many excellent works have been proposed for reconstructing real faces and their expressions as well. Among them, example-based methods first build a low-dimensional parametric representation of 3D face models from an example set and then fit the parametric model to the input 2D image~\cite{blanz1999morphable,cao2014facewarehouse}. The shape-from-shading approach reconstructs faces from image(s) using shading variation~\cite{kemelmacher20113d,kemelmacher20083d}.

Compared to normal face reconstruction, caricature modeling is much more difficult. The challenges lie at least in several aspects outlined below.  First, the diversity of caricatures is much more severe than normal faces, which means example-based methods in real face reconstruction cannot be simply transferred. For example, 3DMM~\cite{blanz1999morphable} and FaceWareHouse~\cite{cao2014facewarehouse} are very successful in modeling normal faces, but the space they define is not large enough for modeling caricatures in our experiments. Second, caricatures are by nature artwork and may not reflect the real physical environment such as lighting information, which implies caricature images may not provide accurate shading cues. Third, creating caricatures is an artistic process. Different caricaturists can develop very different styles for caricatures. Thus the construction process should also consider the individual styles of exaggeration.

Inspired by rapid advances and power of machine learning techniques,  learning-based approaches have also been proposed to create 3D caricature models~\cite{liu2009semi,han2017deepsketch2face}. These approaches require a 3D caricature dataset for training. However, creating a 3D caricature dataset is time-consuming because this usually involves caricaturists and caricatures have richer information such as a large range of deformation possibility and different styles.

Note that caricatures have two basic characteristics. The first one is that they have the face constraint, i.e. we can tell they are faces. The second one is that the features of the face have been exaggerated. These characteristics suggest that a caricature can be viewed as a deformation from a standard face that keeps inherent features of the original face. The effect of exaggeration implies large and nonuniform deformation and usually extrapolation is needed. While classical parametric face models focus on the position of each vertex of 3D faces and they usually use interpolation to create new faces, they have difficulty in producing largely exaggerated faces. We borrow the concept of deformation gradients from mesh deformation and introduce a new deformation representation that is suitable for local and large deformation in a natural way. This representation allows a data-driven approach to generating a deformation space from a normal face dataset, which is flexible and maintains the face-like target. Moreover, we propose to use a set of facial landmarks to capture the exaggeration style of the input 2D caricature image and formulate an optimization problem based on landmark constraints to make sure that the generated 3D caricature has the similar exaggeration style (see Fig.~\ref{fig1}). This avoids the need of creating a 3D caricature dataset with the same style.


The main contributions of the paper are twofold. First, we propose a new intrinsic deformation representation that uses local 
deformation gradients
and allows expressing face-like targets with nonuniform, large local deformation. This deformation representation could also be useful for other applications. Second, we formulate our 3D caricature generation as an optimization problem whose solution delivers a 3D caricature satisfying the face constraint and exaggeration styles. 
\section{Related Work}
Face reconstruction and recognition~\cite{cao2014displaced,guo2017photo,jackson2017large,jiang20173d} are closely relevant to our work. For face reconstruction, data-driven approaches are becoming popular. For example, Blanz and Vetter proposed a 3D morphable model (3DMM)~\cite{blanz1999morphable} that was built on an example set of 200 3D face models describing shapes and textures. Based on 3DMM, Convolutional Neural Networks (CNNs) were constructed to generate 3D face models~\cite{jackson2017large,guo2017photo}.  Cao \etal~\cite{cao2014facewarehouse} used RGBD sensors to develop  FaceWareHouse, a large face database with 150 identities and 47 expressions for each identity.
Using FaceWareHouse, \cite{jiang20173d} regressed the parameters of the bilinear model of \cite{vlasic2005face} to construct 3D faces from a single image. We also use 3D faces in the FaceWareHouse dataset to construct our 3D face representation.

Following Brennan's work~\cite{Brennan85}, many attempts have been made to develop computer-assisted tools or systems for creating 2D caricatures.  Akleman \etal \cite{akleman1997making} developed an interactive tool to make caricatures using morphing. Liang \etal \cite{liang2002example} used a caricature training database and learned exaggeration prototypes from the database using principal component analysis. Chiang \etal \cite{liao2004automatic} developed an automatic caricature generation system by analyzing facial features and using one existing caricature image as the reference. 

Relatively there is much less work on 3D caricature generation~\cite{o1997three,o19993d}. Clarke \etal \cite{clarke2011automatic} proposed an interactive caricaturization system to capture deformation style of 2D hand-drawn caricatures. The method first constructed a 3D head model from an input facial photograph and then preformed deformation for generating 3D caricatures. Liu \etal \cite{liu2009semi} proposed a semi-supervised manifold regularization method to learn a regressive model for mapping between 2D real faces and the enlarged training set with 3D caricatures.
With the development of deep learning, Han \etal \cite{han2017deepsketch2face} developed a sketch system using a CNN to model 3D caricatures from simple sketches. In their approach, the FaceWareHouse~\cite{cao2014facewarehouse} was extended with 3D caricatures to handle the variation of 3D caricature models since the lack of 3D caricature samples made it challenging to train a good model. Different from these works, our approach does not require 3D caricature samples.

In geometric modeling, deformation is a common technique. Many surface based deformation techniques are related to the underlying geometric representation. Local differential coordinates are a powerful representation that encodes local details and can benefit the deformation by preserving shape details~\cite{yu2004mesh,sorkine2005laplacian}. To provide high-level or semantic control, data-driven techniques learn deformation from examples. Sumner \etal~\cite{sumner2004deformation} proposed a deformation method by blending the deformation gradients of example shapes. Baran \etal \cite{baran2009semantic} proposed a semantic deformation transfer method using rotation-invariant coordinates. Gao \etal \cite{gao2016efficient} proposed a deformation representation by blending rotation differences between adjacent vertices and scaling/shear at each vertex and further developed a sparse data-driven deformation method for large rotations~\cite{gao2017sparse}. Our work borrows the concept of local 
deformation gradients to build the deformation representation.
\section{Intrinsic Deformation Representation}

To produce 3D caricatures from 2D images, we first build a new 3D representation for 3D caricature faces. Unlike previous methods that rely on a large set of carefully designed 3D caricature faces for training, our method takes \emph{standard} 3D faces, and exploits the capability of extrapolation of an intrinsic deformation representation. Standard 3D faces are much easier to obtain and are readily available from standard datasets. By contrast, caricature faces are much richer. Particularly, different artists may have different styles for caricatures. Therefore, providing a full coverage is extremely difficult.



\subsection{Deformation representation for 2 models}

To make it easy to follow, we first introduce our intrinsic representation of the deformation between two models, which will then be extended to a collection of shapes.
In particular, one model is chosen as the reference model and the other is the deformed model. We assume they have been globally aligned.
Let us denote by $\mathbf{p}_i$ the position of the $i^{\rm th}$ vertex $v_i$ on the reference model, and by $\mathbf{p}_i'$ the position of $v_i$ on the deformed model. The deformation gradient in the 1-ring neighborhood of $v_i$ from the reference model to the deformed model is defined as the affine transformation matrix $\mathbf{T}_i$ that minimizes the following energy:
\begin{equation}
E(\mathbf{T}_{i}) = \sum_{j\in{\mathcal{N}_i}} c_{ij}\|{{\mathbf{e}_{ij}^{\prime}}-\mathbf{T}_{i}\mathbf{e}_{ij}}\|^2
\end{equation}
where $\mathcal{N}_i$ is the 1-ring neighborhood of vertex $v_i$, $\mathbf{e}_{ij}'=\mathbf{p}_i'-\mathbf{p}_j'$, $\mathbf{e}_{ij}=\mathbf{p}_i-\mathbf{p}_j$, and $c_{ij}$ is the cotangent weight depending only on the reference model to cope with irregular tessellation~\cite{botsch2008linear}. The matrix $\mathbf{T}_i$ can be decomposed into a rotation part $\mathbf{R}_i$ and a scaling/shear part $\mathbf{S}_i$ using polar decomposition: $\mathbf{T}_i=\mathbf{R}_i\mathbf{S}_i$.


To allow effective linear combination, we take the axis-angle representation~\cite{diebel2006representing} to represent the rotation matrix $\mathbf{R}_i$ of vertex $v_i$. The rotation matrix can be represented using a rotation axis $\boldsymbol{\omega}_i$ and rotation angle $\theta_i$ pair with the mapping $\phi$, specifically $\mathbf{R}_i = \phi(\boldsymbol{\omega}_i,\theta_i)$,
where $\boldsymbol{\omega}_i\in {R^3}$ and $\|\boldsymbol{\omega}_i\| =1$. Given two rotations in the axis-angle representation, it is not suitable to blend them linearly, so we convert the axis $\boldsymbol{\omega}_i$ and angle $\theta_i$ to the matrix logarithm representation:
\begin{equation}
\log \mathbf{R}_i = \theta_i
\begin{pmatrix}
0 & -\omega_{i,z} & \omega_{i,y} \\
\omega_{i,z} & 0 & -\omega_{i,x} \\
-\omega_{i,y} & \omega_{i,x} & 0 \\
\end{pmatrix}
\end{equation}

The logarithm of rotation matrices allows effective linear combination~\cite{alexa2002linear}, e.g., two rotations $\mathbf{R}_i$ and $\mathbf{R}_j$ can be blended using
$\exp(\log \mathbf{R}_i+\log \mathbf{R}_j)$. Rotation matrix $\mathbf{R}_i$ can be recovered by matrix exponential $\mathbf{R}_i=\exp(\log \mathbf{R}_i)$.


If the deformed model is the same as the reference model, $\log \mathbf{R}_i=\mathbf{0}$ and $\mathbf{S}_i = \mathbf{I}$ for all $v_i\in{\mathcal{V}}$, where $\mathcal{V}$ is the set of vertices, and $\mathbf{I}$ and $\mathbf{0}$ are identity and zero matrices. Thus we define our deformation representation $\mathbf{f}$ as
\begin{equation}
\mathbf{f}=\{\log \mathbf{R}_i;\mathbf{S}_i-\mathbf{I}| \forall{v_i}\in{\mathcal{V}}\}.
\end{equation}
By subtracting the identity matrix $\mathbf{I}$ from $\mathbf{S}$, the deformation representation of ``no deformation'' becomes a zero vector which builds a natural coordinate system.

Alternative representations~\cite{gao2016efficient,gao2017sparse} mainly focus on representing very large rotations (e.g., $>180^\circ$), which are not needed for faces. In comparison, our representation is simpler and effective.

\subsection{Deformation representation for shape collections}

We can extend the previous definition to a collection of models. Suppose we have $n+1$ 3D face models. In our experiments,  we select face models from \emph{FaceWareHouse}~\cite{cao2014facewarehouse}, a 3D face dataset with large
variation in identity and expression. At first, we mark some facial landmarks on the 3D face models. With the help of the landmarks, we apply rigid alignment to 3D face models to remove global rigid transformation between them.

We similarly choose one model as the reference model and let the others be the deformed models. Given $n$ deformed models, we can obtain $n$ deformation representations $\mathcal{F}=\{\mathbf{f}_l | l =1,\cdots,n\}$ with
\begin{equation}
\mathbf{f}_l=\{\log \mathbf{R}_{l,i};\mathbf{S}'_{l,i}| \forall{v_i}\in{\mathcal{V}}\}
\end{equation}
and $\mathbf{S}'_{l,i}=\mathbf{S}_{l,i}-\mathbf{I}$ for simpler expression. 

\begin{figure}[t]
\centering
   \includegraphics[width=1\linewidth]{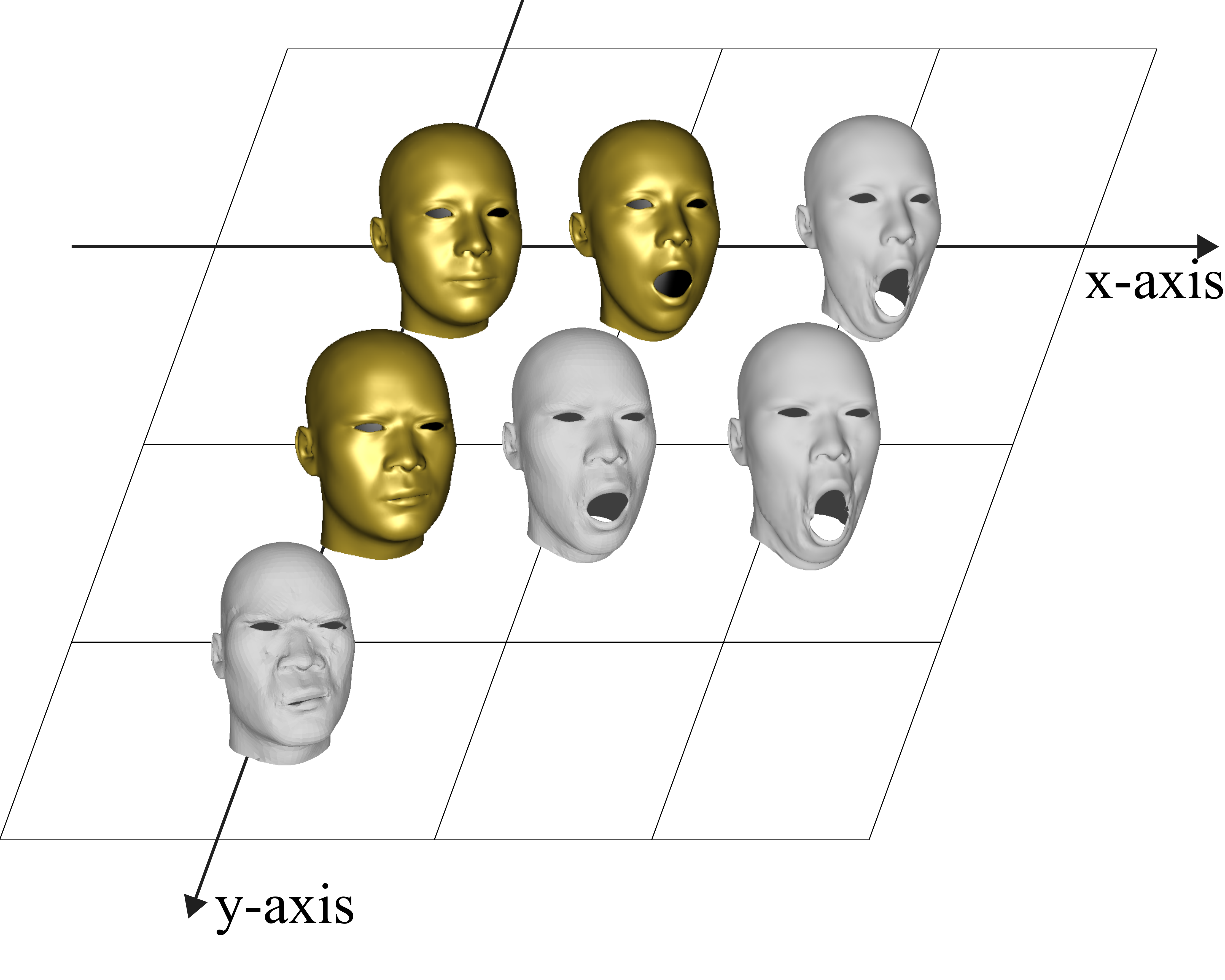}
   \caption{A simple example of the deformation representation coordinate system. Here, the reference model is at the origin, and two example shapes are placed at $(1,0)$ (with an open mouth) and $(0, 1)$ (with a different identity). Therefore, the $x$ and $y$ axes denote to these two modes of deformation.  Denote by $\mathbf{f}_1$ and $\mathbf{f}_2$ the two basis deformations, we
can obtain new 3D faces using a  linear combination of them, including exaggerated shapes. }
\label{fig2}
\setlength{\belowcaptionskip}{-1cm}
\end{figure}

The deformation representation $\mathcal{F}$ actually defines a deformation space. To generate a new deformed mesh based on $\mathcal{F}$, we formulate the deformation gradient of a deformed mesh as a linear combination of the basis deformations $\mathbf{f}_l$:
\begin{equation}\label{eqn:Ti}
\mathbf{T}_i(\mathbf{w}) = \exp(\sum_{l=1}^n{w_{R,l}}\log \mathbf{R}_{l,i}){(\mathbf{I} + \sum_{l=1}^nw_{S,l} \mathbf{S}'_{l,i})}
\end{equation}
where $\mathbf{w}=(\mathbf{w}_R, \mathbf{w}_S)$ is the combination weight vector, consisting of weights of rotation $\mathbf{w}_R=\{w_{R,l}|l =1,\cdots,n\}$ and weights of scaling/shear $\mathbf{w}_S=\{w_{S,l}|l =1,\cdots,n\}$. $\log \mathbf{R}_{l,i}$ and $\mathbf{S}'_{l.i}$ correspond to the rotation and scaling/shear of the $i^{\rm th}$ vertex in the $l^{\rm th}$ basis $\mathbf{f}_l$.
Given the representation basis $\mathcal{F}$, different faces can be obtained by varying $\mathbf{w}$. By introducing two sets of weights for rotation and scaling/shear, our representation is flexible to cope with exaggerated 3D faces, as we will demonstrate later.

We give a simple example in Fig.~\ref{fig2} where $\mathbf{w}_R=\mathbf{w}_S$ (and denoted as $\mathbf{w}$ for simplicity).
We have two deformed example models, so $\mathbf{w}$ only has 2 dimensions. The golden face located in the origin is the reference model. Two deformed models are shaded golden and located at $(1, 0)$ (with the mouth open), and $(0, 1)$ (with a different subject).
By setting $\mathbf{w}$ to $(2,0)$, we can let the mouth open wider.
If we set $\mathbf{w}$ to $(1,1)$, we can let the face at $(0,1)$ open his mouth. When $\mathbf{w}$ is set to $(0,2)$, it will exaggerate the face at $(0,1)$. By a linear combination of deformation basis $\mathcal{F}$ using different weights $\mathbf{w}$, we can generate deformed meshes in the deformation space.

\subsection{Deformation weight extraction and model reconstruction}

We can define the deformation energy $E_{def}$ as follows:
\begin{equation}\label{eqn:Edef}
E_{def}(\mathbf{P}', \mathbf{w})= \sum_{v_i\in{\mathcal{V}}}\sum_{j\in{\mathcal{N}_i}} c_{ij}\|{({\mathbf{p}_i^{\prime}}-{\mathbf{p}_j^{\prime}})-\mathbf{T}_{i}(\mathbf{w})(\mathbf{p}_i-\mathbf{p}_j)}\|^2
\end{equation}
where $\mathbf{P'} = \{\mathbf{p}_i'| v_i \in \mathcal{V}\}$ represents the positions of deformed vertices. By minimizing this energy, we are able to determine the position of each vertex $\mathbf{p}_i' \in \mathbb{R}^3$ on the deformed mesh given weights $\mathbf{w}$, or obtain the combination weights $\mathbf{w}$ given the deformed mesh $\mathbf{P}'$.

\subsubsection{Model reconstruction from weights $\mathbf{w}$} \label{sec:recons}

Given $\mathbf{w}$, the deformation gradient $\mathbf{T}(\mathbf{w})$ can be directly obtained using Eq.~\ref{eqn:Ti}. Then model reconstruction is done by finding the optimal $\mathbf{P}'$ that minimizes:
\begin{equation}
E_{def}(\mathbf{P}')= \sum_{v_i\in{\mathcal{V}}}\sum_{j\in{\mathcal{N}_i}} c_{ij}\|{({\mathbf{p}_i^{\prime}}-{\mathbf{p}_j^{\prime}})-\mathbf{T}_{i}(\mathbf{w})(\mathbf{p}_i-\mathbf{p}_j)}\|^2.
\end{equation}
For each ${\mathbf{p}_i^{\prime}}$, we tackle it by solving $\frac{\partial{E_{def}(\mathbf{P}')}}{\partial \mathbf{p}_i'} =0$, 
which leads to:
\begin{equation}
2\sum_{j\in{\mathcal{N}_i}} c_{ij}\mathbf{e}_{ij}' =\\
\sum_{j\in{\mathcal{N}_i}} c_{ij}(\mathbf{T}_i(\mathbf{w}) + \mathbf{T}_j(\mathbf{w}))\mathbf{e}_{ij}
\end{equation}
with $\mathbf{e}_{ij}=\mathbf{p}_i -\mathbf{p}_j$, $\mathbf{e}'_{ij}=\mathbf{p}'_i -\mathbf{p}'_j$. The resulting linear system can be written in the form $\mathbf{Ap}'=\mathbf{b}$ where the matrix $\mathbf{A}$ is fixed and sparse since only entries where the corresponding vertices are associated with the edge are non-zero. By specifying the position of one vertex, we can get single solution to the equations. This initial specification will not change the shape of the output. The models shown in Fig.~\ref{fig2} are obtained using this optimization.

\subsubsection{Optimizing $\bm{w}$ for a given deformed model}\label{sec:w}

Given a deformed 3D model $\mathbf{P}'$, the optimal weights $\mathbf{w}$ to represent the deformation can be obtained by minimizing:
\begin{equation}
E_{def}(\mathbf{w})= \sum_{v_i\in{\mathcal{V}}}\sum_{j\in{\mathcal{N}_i}} c_{ij}\|{({\mathbf{p}_i^{\prime}}-{\mathbf{p}_j^{\prime}})-\mathbf{T}_{i}(\mathbf{w})(\mathbf{p}_i-\mathbf{p}_j)}\|^2.
\end{equation}
This is a non-linear least squares problem because of $\mathbf{T}_i(\mathbf{w})$. To solve it, we first compute Jacobian matrix  $\frac{\partial \mathbf{T}_i(\mathbf{w})}{\partial w_{l}}$ w.r.t. example model $l$ derived as two components:
\begin{equation}
\begin{cases}
\frac{\partial \mathbf{T}_i(\mathbf{w})}{\partial w_{R,l}} =
\exp({\sum_l{w_{R,l} \log \mathbf{R}_{l,i}})\log \mathbf{R}_{l,i}(I+\sum_lw_{S,l}\mathbf{S}'_{l,i})}\\
\frac{\partial \mathbf{T}_i(\mathbf{w})}{\partial w_{S,l}} =
\exp({\sum_l{w_{R,l} \log \mathbf{R}_{l,i}}})\mathbf{S}'_{l,i}
\end{cases}
\end{equation}
which are the derivatives of $\mathbf{T}_i$ w.r.t. the rotation weight $\mathbf{w}_R$ and scaling/shear $\mathbf{w}_S$, respectively.
The optimal $\mathbf{w}$ for a given deformed model can be calculated using the \emph{Levenberg-Marquardt} algorithm~\cite{more1978levenberg}.

\section{Generation of 3D Caricature Models}

Built upon our deformation representation, we now describe our algorithm to construct a 3D caricature model from a 2D caricature image. Assume that we have already had a 3D reference face model and a deformation representation based on it.
To capture exaggerated facial expressions, we use a set of landmark points in the 2D image and 3D model, which correspond to the landmark points marked on the reference face.


Reconstructing the 3D model from a 2D image is the inverse process of observing a 3D object by projecting it to an imaging plane. Therefore, this process is affected by view parameters. For simplicity, we  choose orthographic projection to set the relationship between 3D and 2D. Without loss of generality, we assume that the projection plane is the $z$-plane and thus the projection can be written as
\begin{equation}
\mathbf{q}_i = s
\begin{pmatrix}
1 & 0 & 0 \\
0 & 1 & 0
\end{pmatrix}
\mathbf{Rp}_i+\mathbf{t}
\end{equation}
where $\mathbf{p}_i$ and $\mathbf{q}_i$ are the locations of vertex $v_i$ in the world coordinate system and in the image plane, respectively, $s$ is the scale factor, $\mathbf{R}$ is the rotation matrix constructed from Euler angles \emph{pitch, yaw, roll}, and $\mathbf{t}=(t_x,t_y)^T$ is the translation vector. For convenience we introduce $\mathbf{\Pi}$ as:
\begin{equation}
\mathbf{\Pi} =
s\begin{pmatrix}
1 & 0 & 0 \\
0 & 1 & 0
\end{pmatrix}.
\end{equation}
Then we map 3D landmarks onto the image plane by $\Pi \mathbf{R} \mathbf{p}_i +\mathbf{t}$. The landmark fitting loss can be defined as
\begin{equation}\label{eqn:Elan}
E_{lan}(\mathbf{\Pi,R,t,\mathcal{L}})=\sum_{v_i \in \mathcal{L}}\|\Pi \mathbf{R} \mathbf{p}_i' +\mathbf{t}-\mathbf{q}_i\|^2
\end{equation}
where $\mathcal{L}$ and $\mathcal{Q}=\{\mathbf{q}_i|v_i \in \mathcal{L}\}$ are the set of 3D landmarks and 2D landmarks.

To generate a 3D caricature model that looks like a human face and meanwhile matches 2D landmarks for the effects of exaggeration, we utilize
our deformation representation and the projection relationship. The problem is formulated as an optimization problem:
\begin{equation}\label{eq:optimization_caricature}
\min_{\mathbf{P'},\mathbf{w}} E_{def}(\mathbf{P'},\mathbf{w}) + \lambda E_{lan}(\mathbf{\Pi,R,t},\mathcal{L})
\end{equation}
where $E_{def}$ and $E_{lan}$ are defined in Eq.~\ref{eqn:Edef} and Eq.~\ref{eqn:Elan}, and  $\lambda$ is the tradeoff factor controlling the relative importance of the two terms.


To solve the above optimization problem, we initialize $\mathbf{w}$ by simply letting all weights be zero and then alternately solve for $\mathbf{P}'$ and $\mathbf{w}$ using the following $\mathbf{P}'$-step and $\mathbf{w}$-step. The process continues until convergence or reaching the maximum number of iterations. The whole algorithm is outlined in Algorithm~\ref{alg:Framwork}.

\noindent
\textbf{$\mathbf{P'}$-step:} We use the similar approach described in Sec.~\ref{sec:recons} to obtain $\mathbf{P}'$. Let $E$ be the overall energy to be optimized. We set $\frac{\partial E}{\partial \mathbf{p}'_i} = 0$ which gives the following equations:
\begin{equation}
\begin{cases}
\begin{split}
2\sum_{j\in{\mathcal{N}_i}} c_{ij}\mathbf{e}_{ij}' +&\lambda \mathbf{R}^{T} \mathbf{\Pi}^{T} \mathbf{\Pi}\mathbf{R} \mathbf{p}_i' =\sum_{j\in{\mathcal{N}_i}} c_{ij}\mathbf{T}_{ij}(\mathbf{w})\mathbf{e}_{ij}\\
& + \lambda \mathbf{R}^{T} \mathbf{\Pi}^{T}(\mathbf{q}_i-\mathbf{t}), (v_i \in \mathcal{L})
\end{split}   \\
2\sum_{j\in{\mathcal{N}_i}} c_{ij}\mathbf{e}_{ij}' =
\sum_{j\in{\mathcal{N}_i}} c_{ij}\mathbf{T}_{ij}(\mathbf{w})\mathbf{e}_{ij}, (v_i \notin \mathcal{L}) 
\end{cases}
\end{equation}
where $\mathbf{T}_{ij}(\mathbf{w})=\mathbf{T}_i(\mathbf{w})+ \mathbf{T}_j(\mathbf{w})$. The former equation applies to landmark vertices ($v_i \in \mathcal{L}$) and the latter equation applies to non-landmark vertices ($v_i \notin \mathcal{L}$). 

\noindent
\textbf{$\mathbf{w}$-step:} In this step, we fix $\mathbf{P}'$ and optimize $\mathbf{w}$. Since $E_{lan}$ is independent of $\mathbf{w}$, we use the  \emph{Levenberg-Marquardt} algorithm described in Sec.~\ref{sec:w} to solve the non-linear least squares problem. After solving for $\mathbf{w}$, we first update projection parameters then go back to optimize $\mathbf{P'}$-step. We exit the loop to obtain the generated $\mathbf{P}'$.

\begin{algorithm}[htb]
\caption{ Generation of 3D caricature models}
\label{alg:Framwork}
\begin{algorithmic}
\REQUIRE ~~\\ 
Caricature image $I$; \\
A reference 3D face; \\
Deformation representation $\mathcal{F}=\{\mathbf{f}_l\}$;
\ENSURE 3D caricature model mesh vertex positions $\mathbf{P}'$\\ 
\STATE {Generate 2D landmarks $\{\mathbf{q}_i\}$}
\STATE {Initialize $\mathbf{w}$}
\FOR{each iteration}
\STATE {update projection parameters $\mathbf{\Pi},\mathbf{R,t}$}
\STATE {Solve for $\mathbf{P}'$ in the $\mathbf{P}'$-step}
\IF {$\Delta E_{def}<\epsilon$ or $reach \ max \ iteration\ times$}  
\STATE exit;
\ELSE
\STATE Solve for $\mathbf{w}$ in the $\mathbf{w}$-step
\ENDIF
\ENDFOR
\end{algorithmic}
\end{algorithm}

\section{Experiments}
\subsection{Implementation Details}

\begin{figure}
\begin{center}
   \includegraphics[width=1\linewidth]{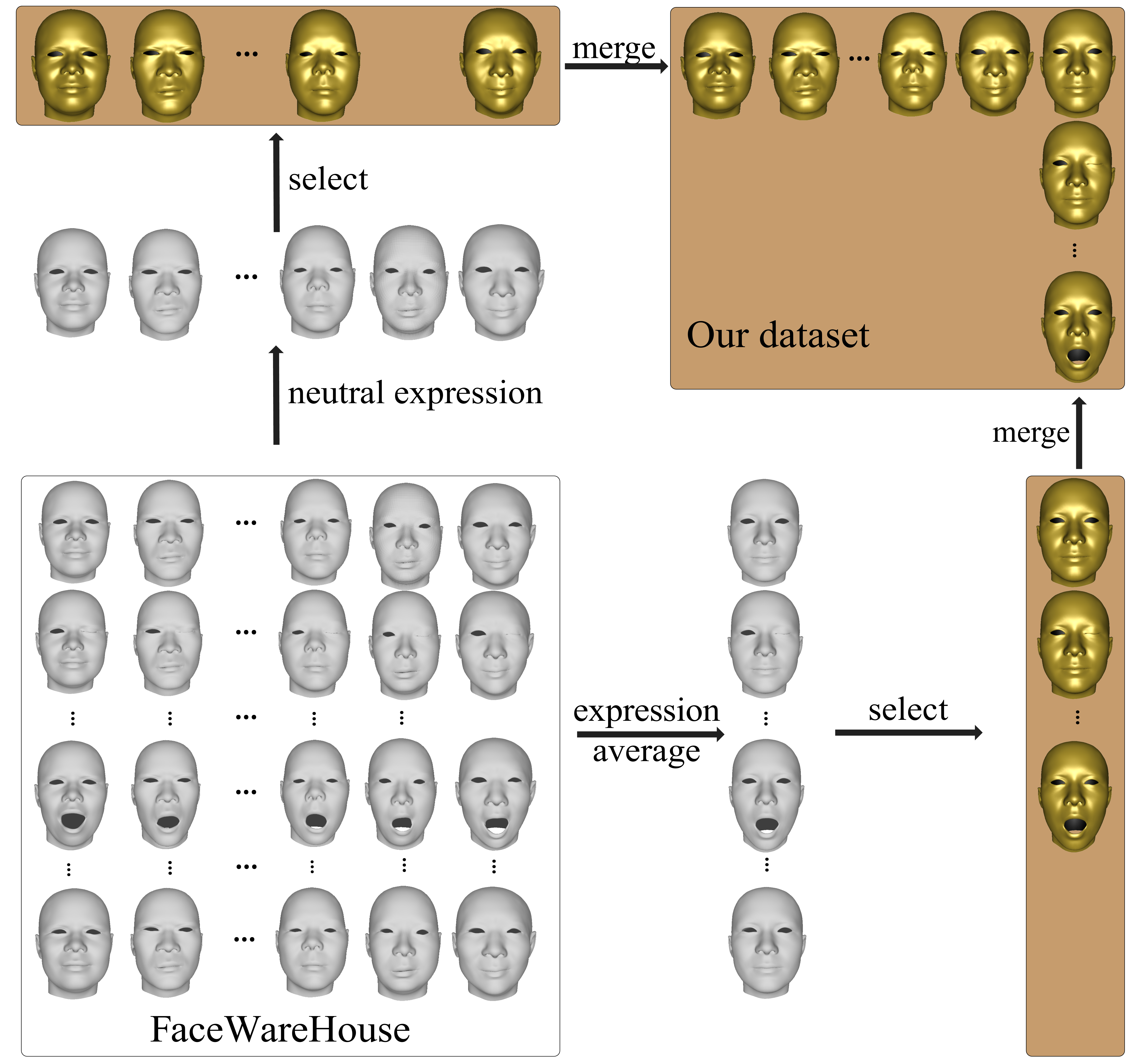}
\end{center}
   \caption{The pipeline of our database construction. We generate the average face for each expression and select 23 expressions of average faces. Meanwhile we choose 75 identities with neutral expression. Merging both gives our dataset.}
\label{dataset}
\end{figure}

Our algorithm is implemented in C++, using the Eigen library~\cite{eigenweb} for all linear algebra operations. All the examples are run on a desktop PC with 4GB of RAM and a hexa-core CPU at 1.6 GHz. We set $\lambda$ in Eq.~\ref{eq:optimization_caricature} to be 0.01 and the initial value of $\mathbf{w}$ is set to be a zero vector. For all the test models, we run our solver for 4 iterations in Alg.~\ref{alg:Framwork} and $\epsilon = 10^{-2}$, which are sufficient to get satisfactory results. The number of vertices is 11510. The average time for manually adjusting the landmarks is about $2\sim 4$ min. To solve the optimization problem, each iteration takes about $7\sim 10$s including $\mathbf{P'}$-step and $\mathbf{w}$-step where $\mathbf{w}$-step takes most of the time. Overall, it takes less than 40s to produce the result with our unoptimized implementation.

To construct the deformation representations $\mathcal{F}=\{\mathbf{f}_l|l\in{\{1,...,n\}}\}$, we choose models from FaceWareHouse dataset~\cite{cao2014facewarehouse}, which contains 150 identities and 47 expressions for each identity. The construction pipeline is shown in Fig.~\ref{dataset}. We first compute the average shape of each expression and then select $23$ expressions with large differences. Meanwhile we choose neutral expression of each identity and select $75$ models with large differences to the mean neutral face. Merging these two parts together, we obtain our dataset, which includes $98$ face models. The mean neutral face is set as the reference model and the others are set as deformed models.

\begin{figure}[t]
\begin{center}
   \includegraphics[width=1\linewidth]{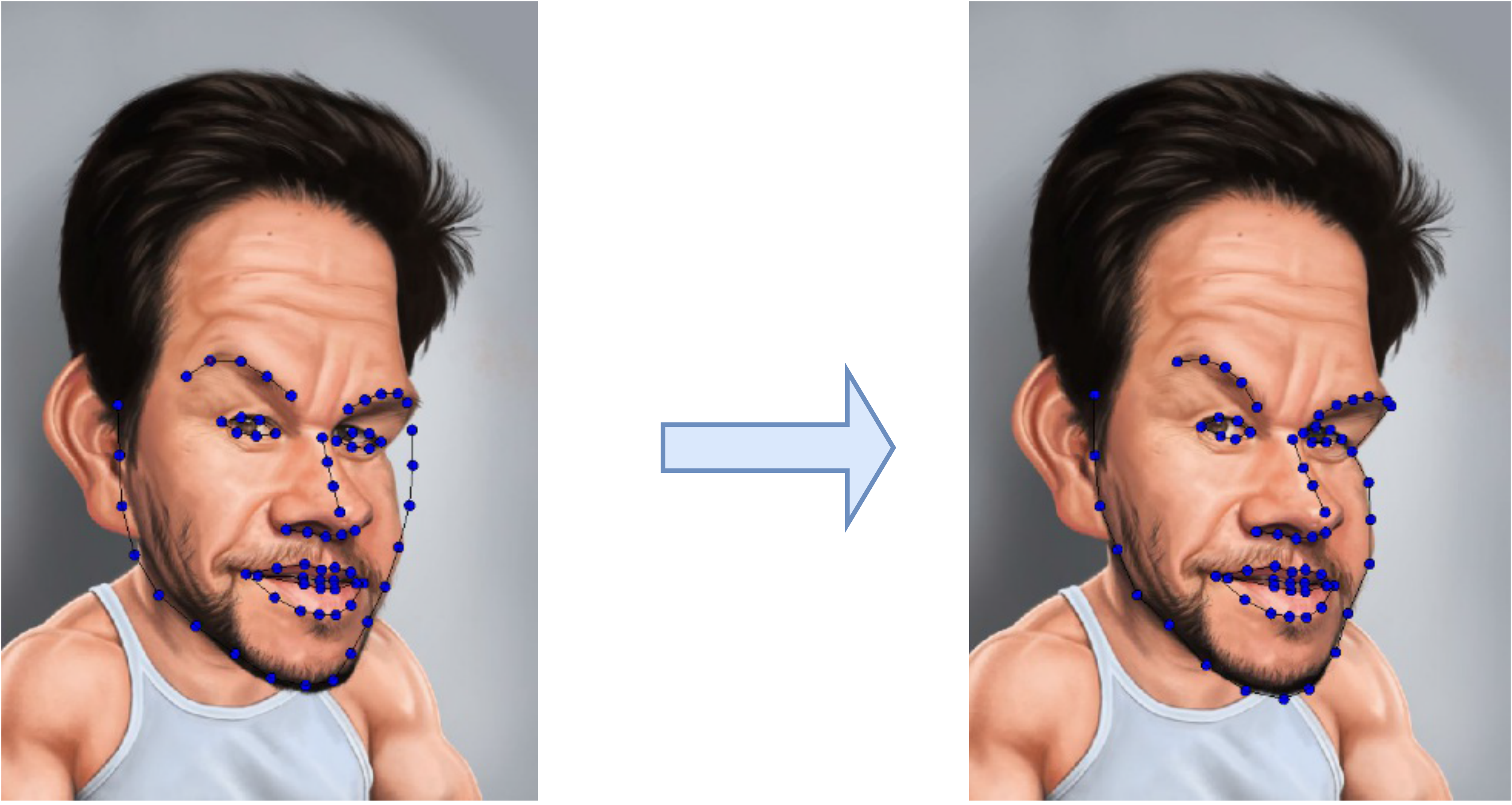}
\end{center}
   \caption{Some of facial landmarks on the 2D caricature image detected by Dlib may not be accurate (left). We design an interactive system to allow the user to relocate the landmarks (right).}
\label{landmark}
\end{figure}

In our method, 68 facial landmarks are applied to constrain the 3D face shape. To detect the facial landmarks on caricature images, we use Dlib library~\cite{king2009dlib}. However, as Dlib library is trained with \emph{standard} face images, some of the detected landmarks on caricature images might be inaccurate. We design an interactive system that allows the user to adjust the landmarks, as shown in Fig.~\ref{landmark}.


In Alg.~\ref{alg:Framwork}, we update $\mathbf{\Pi,R,t}$ before $\mathbf{P'}$-step. Similar with ~\cite{jiang20173d}, we update the silhouette landmark vertices for non-frontal caricature according to the rotation matrix $\mathbf{R}$. The projection parameters are updated via a linear least squares optimization by fixing $\mathcal{L}$ and $\mathcal{Q}$ in Eq.~\eqref{eqn:Elan}.

\begin{figure}
\begin{center}
   \includegraphics[width=1\linewidth]{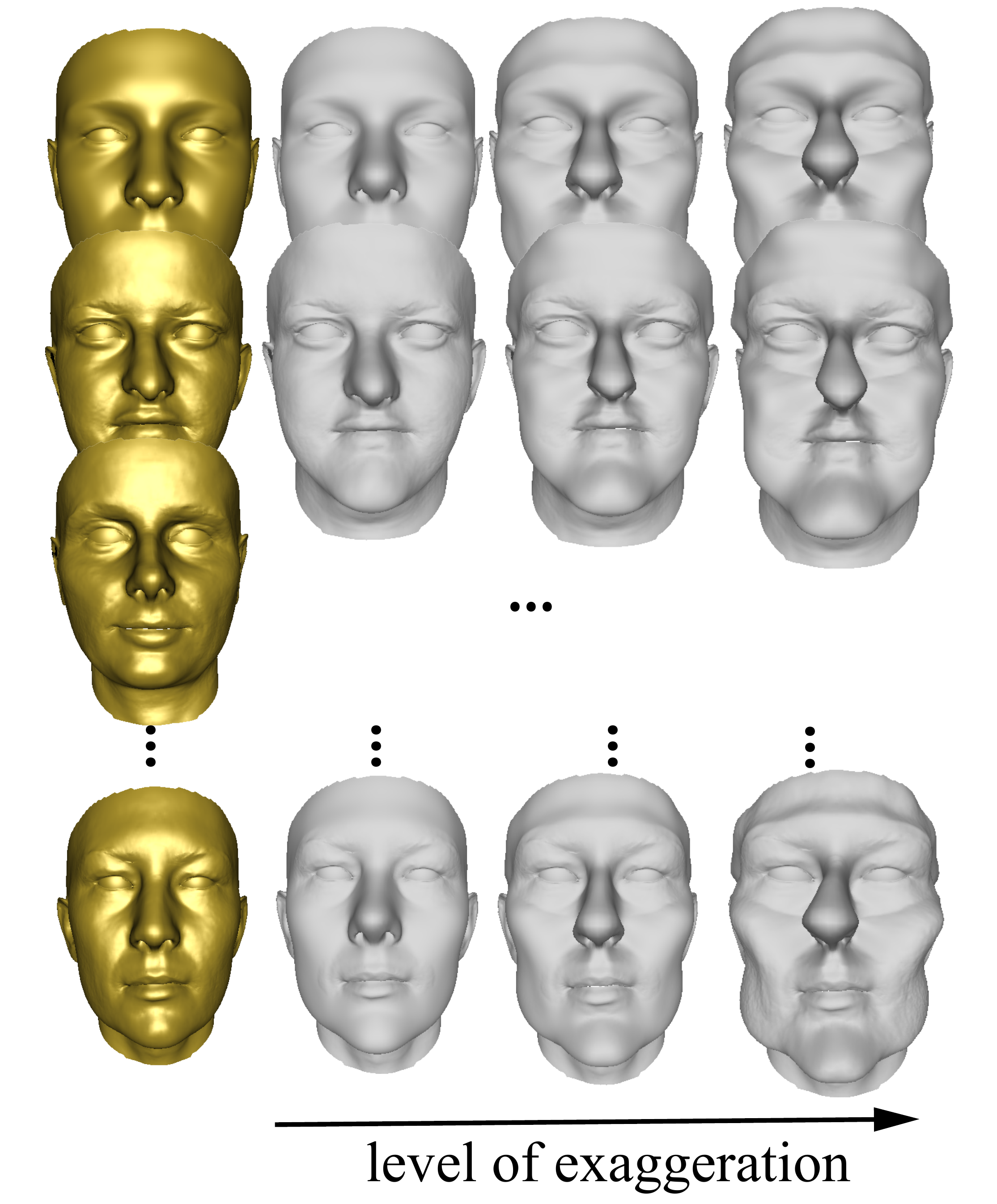}
\end{center}
   \caption{We follow the method in~\cite{han2017deepsketch2face,sela2015computational} to expand the Basel Face Model~\cite{paysan20093d} to generate some caricatured face models. Golden faces in the first column are the original models from the dataset and we generate three levels of caricatured models.}
\label{3dmm_cari}
\end{figure}

\begin{figure*}
\begin{center}
   \includegraphics[width=1\linewidth]{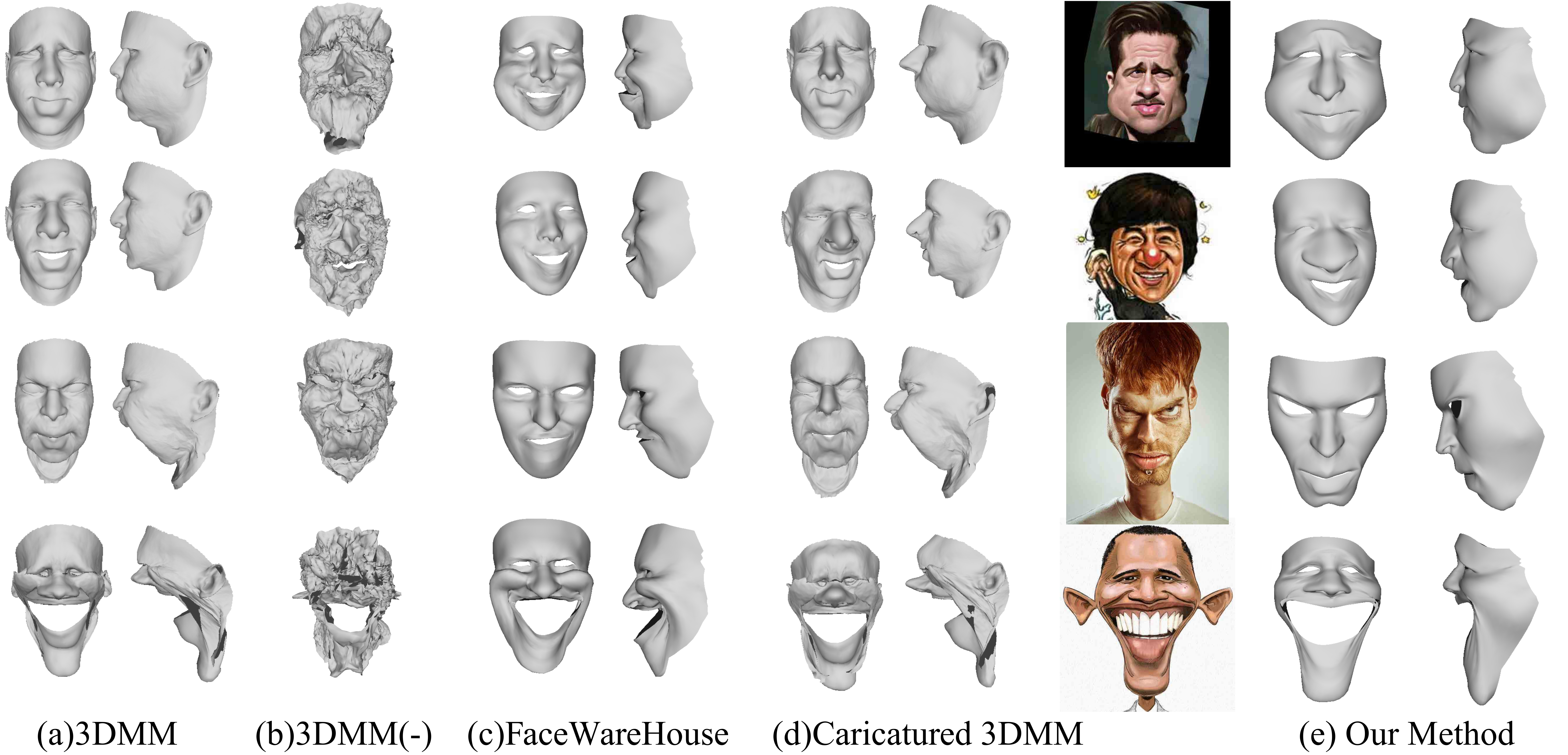}
\end{center}
   \caption{Comparison among (a) 3DMM~\cite{zhu2015high}, (b) 3DMM(-), (c) FaceWareHouse~\cite{cao2014facewarehouse,cao20133d,cao2014displaced}, (d) Caricatured 3DMM~\cite{han2017deepsketch2face,sela2015computational} and (e) our method. Except 3DMM(-), we show generated models with front and right views.}
\label{fig:compared}
\end{figure*}

\subsection{Baselines}
We compare the proposed method with the reconstruction methods by 3DMM~\cite{paysan20093d,zhu2015high} and FaceWareHouse dataset~\cite{cao2014facewarehouse,cao2014displaced}. For all the methods, the 3D face model is reconstructed by minimizing the residuals between the projected 3D landmarks and the corresponding 2D landmarks. In the following, we introduce the implementation details of these baseline methods.

\textbf{3DMM with different regularization weights:} In~\cite{zhu2015high},  the 3DMM representation is adopted to represent any 3D face model:
\begin{equation}
\mathbf{P}=\bar{\mathbf{P}}+\mathbf{A}_{id}\boldsymbol{\alpha}_{id}+\mathbf{A}_{exp}\boldsymbol{\alpha}_{exp},
\end{equation}
where $\mathbf{P}$ is a 3D face, $\bar{\mathbf{P}}$ is the mean face, and $\mathbf{A}_{id}$ and $\mathbf{A}_{exp}$ are the principal axes on identities and expressions. Since $\mathbf{A}_{id}$ and $\mathbf{A}_{exp}$ are generated by principal component analysis, the fitted model should satisfy the distribution of face space, and thus a regularization term is added:
\begin{equation}
E_{reg}(\boldsymbol{\alpha}) = \sum_{j=1}^{199}({\frac{\mathbf{\alpha}_{id,j}}{\mathbf{\sigma}_{id,j}}})^2+\sum_{j=1}^{29}({\frac{\mathbf{\alpha}_{exp,j}}{\mathbf{\sigma}_{exp,j}}})^2,
\end{equation}
where $\mathbf{\sigma}_{id}$ and $\mathbf{\sigma}_{exp}$ are the standard deviations of each principal component for identity and expression, respectively. By representing projected 3D landmarks $\mathcal{L}$ with $\boldsymbol{\alpha}$, the objective energy functional in their method is:
\begin{equation}
\min_{\mathbf{\Pi,R,t},\boldsymbol{\alpha}} E_{lan}(\mathbf{\Pi,R,t},\boldsymbol{\alpha}) + \lambda_{reg}E_{reg}(\boldsymbol{\alpha}).
\label{eqn:3dmmenergy}
\end{equation}
When $\lambda_{reg}$ gets larger, the reconstructed model would get towards mean face. We test $\lambda_{reg}$ with two values $3000$ and $0$, and denote the corresponding reconstruction results as 3DMM, 3DMM(-) respectively.

\textbf{FaceWareHouse:} In~\cite{cao20133d,cao2014displaced,cao2014facewarehouse}, the bilinear model represents 3D face as:
\begin{equation}
\mathbf{P}=C_r\times_2 \mathbf{u}_{id}\times_3 \mathbf{u}_{exp},
\end{equation}
where $C_r$ is the core tensor, and $\mathbf{u}_{id}$ and $\mathbf{u}_{exp}$ are the coefficients of identity and expression.
By minimizing the residuals between projected landmarks and 2D landmarks, we can obtain optimal $\mathbf{u}_{id}$ and $\mathbf{u}_{exp}$ and thus the reconstructed face model.

\textbf{Caricatured 3DMM:} In~\cite{han2017deepsketch2face}, the FaceWareHouse dataset is expanded by adding more exaggerated face models to enhance the representation capability of the original dataset, where the exaggerated face models are generated via the method in~\cite{sela2015computational}. However, the expanded FaceWareHouse dataset is not publicly available. Thus, we follow the same method~\cite{sela2015computational} to expand \emph{Basel Face Model} ~\cite{paysan20093d} dataset, and some examples are shown in Fig.~\ref{3dmm_cari}. After constructing expanded 3DMM, we apply principal component analysis to generate the parametric model, similar to the method in~\cite{han2017deepsketch2face} to produce the bilinear model.

\begin{figure*}
\begin{center}
   \includegraphics[width=1.0\linewidth]{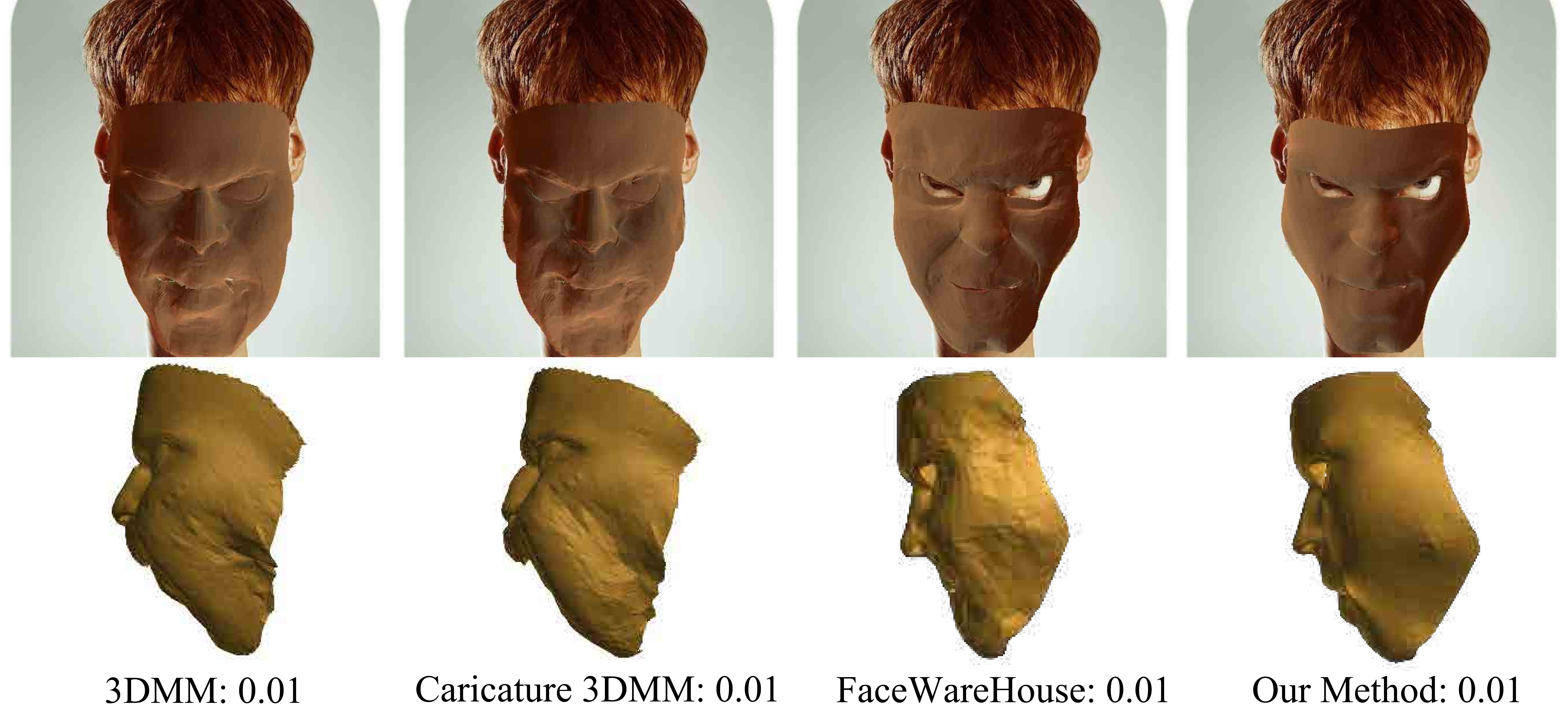}
\end{center}
   \caption{Comparison of 3D caricature results with ARAP deformation (two views per column) of Example 3 in Fig.~\ref{fig:compared}.} 
   \label{fig:rebuttal}
\end{figure*}

\subsection{Results}

\begin{figure*}
\begin{center}
   \includegraphics[width=1\linewidth]{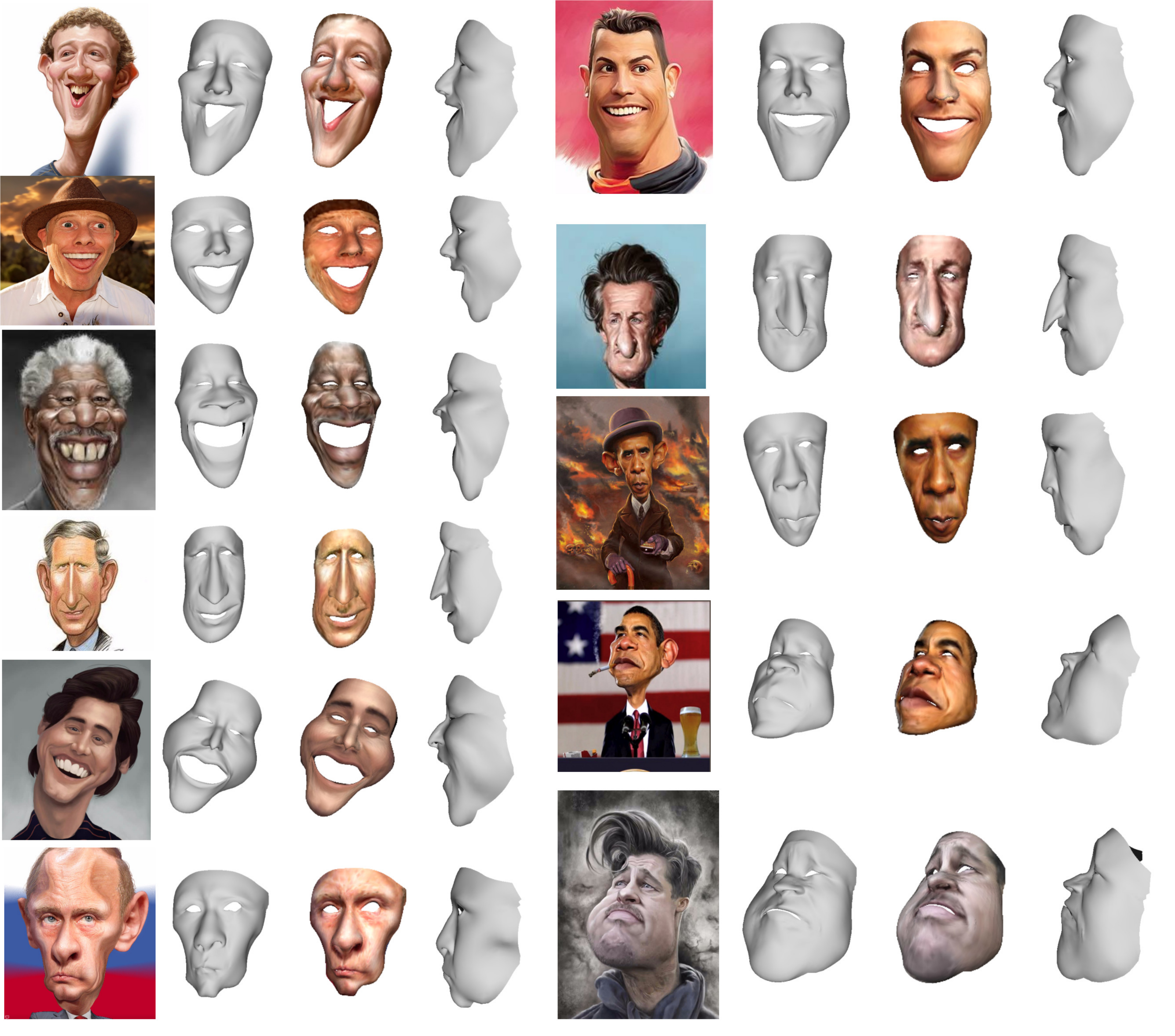}
\end{center}
   \caption{A gallery of 3D caricature models generated by our method.}
\label{fig:model_show}
\end{figure*}

Fig.~\ref{fig:compared} shows some visual results of the generated 3D caricatures using different methods. It can be observed that the face shape by 3DMM is too regular, and thus not exaggerated enough to match the input face images. Although the 3D models by 3DMM(-) are exaggerated, the shapes are distorted, not regular to be face shapes. Since the linear model is applied in FaceWareHouse, its exaggeration capability is also limited. As for the method of caricatured 3DMM, although some caricatured face models generated by using~\cite{sela2015computational} are included in the database, it contains only limited styles of exaggeration. If the exaggerated face is beyond the expanded database, it cannot be properly expressed. In contrast, the reconstruction results by our method are quite close to the shape of the input images, and the model quality is better.

To quantitatively compare different methods, we define the average fitting error $E_{error}$ as the root-mean-square fitting error:
\begin{equation}
E_{error} = \sqrt{E_{lan}/68}.
\end{equation}
We compared 3DMM, 3DMM(-), FaceWareHouse, Caricatured 3DMM and our method over all test data including 50 annotated caricature images, and the statistics are shown in Tab.~\ref{table:fitting}. It can be observed that existing linear 3D face representation methods cannot fit the landmarks of the caricature images well even by setting $\lambda_{reg} = 0$, while our method can achieve small fitting errors.

\begin{table}
\caption{The mean square fitting error of landmarks over test data. The first row shows the methods, and the second row shows their corresponding mean square fitting error of landmarks. FWH: FaceWareHouse; C-3DMM: Caricatured 3DMM.}
\centering
\begin{tabular}{|c|c|c|c|c|}
\hline
3DMM&3DMM(-)&FWH&C-3DMM&Ours\\
\hline
$7.24$&$4.86$&$19.00$&$6.46$&$0.03$ \\
\hline
\end{tabular}
\label{table:fitting}
\end{table}

To further satisfy the landmark constraints while preserving a reasonable face shape, we apply as-rid-as-possible (ARAP)~\cite{sorkine2007rigid} deformation to the reconstruction results of existing methods. With the help of ARAP deformation, the fitting error $E_{error}$ of other methods are reduced to 0.01. Although ARAP deformation could help fit the landmarks, it still preserves the shape structure of the original reconstruction results which cannot fit the input caricature images well, as shown in Fig.~\ref{fig:rebuttal}.

As the landmark fitting error is not sufficient for evaluation, we conduct a user study including 15 participants from different backgrounds. Each participant is given 10 randomly selected images and their corresponding reconstruction results by each method (Our method and ARAP deformation on 3DMM, Caricature 3DMM and FaceWareHouse) in random order. Each participant sorts the reconstructed models from best to worst by judging their similarities with the input caricature image. The statistical result indicates that our models get $80\%$ voting for top-1 and $100\%$ for top-2. We also investigate the subjective feeling of each participant after knowing the corresponding method of each model. According to the following investigation, all participants agree that our method captures the shape of caricatures much better than other three methods. However, in around $20\%$ cases, the output meshes by our method are too smooth compared with other methods, and seem to lack detail information of caricatures. Our reconstructed meshes may contain self-intersections if the expression is exaggerated too much. In comparison, the reconstructed meshes by other methods have the same problem even for caricature images with small exaggerated expressions.

More results by our method are shown in Fig.~\ref{fig:model_show}, where the caricature images are selected from the database~\cite{huo2017webcaricature} and Internet. All the tested caricature images, their corresponding landmarks and the reconstructed meshes are available at \url{https://github.com/QianyiWu/Caricature-Data}.

\section{Conclusion}
We have presented an efficient algorithm for generating a 3D caricature model from a 2D caricature image. As we can see from the experiments, our algorithm can generate appealing 3D caricatures with the similar exaggeration style conveyed in the 2D caricature images. One of the key techniques of our approach is a new deformation representation which has the capacity of modeling caricature faces in a non-linear and more nature way. As a result, compared to previous work, our approach has a unique advantage that we just use normal face models to create caricatures. This enables our approach to be more suitable for various applications in real scenarios.




\noindent \textbf{Acknowledgement} 
We thank Luo Jiang, Boyi Jiang, Yudong Guo, Wanquan Feng, Yue Peng, Liyang Liu and Kaiyu Guo for their helpful discussion during this work. We also thank Thomas Vetter \etal and Kun Zhou \etal for allowing us to use their 3D face datasets. We are grateful to all the artists who created caricatures used in this work and all the participants in the user study. This work was supported by the National Key R\&D Program of China (No. 2016YFC0800501), the National Natural Science Foundation of China (No. 61672481), NTU CoE Grant and MOE Tier-2 Grants (2016-T2-2-065, 2017-T2-1-076). 

{\small
\bibliographystyle{ieee}
\bibliography{egbib}

\begin{thebibliography}{10}\itemsep=-1pt

\bibitem{akleman1997making}
E.~Akleman.
\newblock Making caricatures with morphing.
\newblock In {\em ACM SIGGRAPH}, page 145, 1997.

\bibitem{alexa2002linear}
M.~Alexa.
\newblock Linear combination of transformations.
\newblock In {\em ACM Transactions on Graphics}, volume~21, pages 380--387,
  2002.

\bibitem{baran2009semantic}
I.~Baran, D.~Vlasic, E.~Grinspun, and J.~Popovi{\'c}.
\newblock Semantic deformation transfer.
\newblock {\em ACM Transactions on Graphics}, 28(3):36, 2009.

\bibitem{blanz1999morphable}
V.~Blanz and T.~Vetter.
\newblock A morphable model for the synthesis of {3D} faces.
\newblock In {\em ACM SIGGRAPH}, pages 187--194, 1999.

\bibitem{botsch2008linear}
M.~Botsch and O.~Sorkine.
\newblock On linear variational surface deformation methods.
\newblock {\em IEEE Transactions on Visualization and Computer Graphics},
  14(1):213--230, 2008.

\bibitem{Brennan85}
S.~Brennan.
\newblock The dynamic exaggeration of faces by computer.
\newblock {\em Leonardo}, 18(3):170 -- 178, 1985.

\bibitem{cao2014displaced}
C.~Cao, Q.~Hou, and K.~Zhou.
\newblock Displaced dynamic expression regression for real-time facial tracking
  and animation.
\newblock {\em ACM Transactions on Graphics}, 33(4):43, 2014.

\bibitem{cao20133d}
C.~Cao, Y.~Weng, S.~Lin, and K.~Zhou.
\newblock {3D} shape regression for real-time facial animation.
\newblock {\em ACM Transactions on Graphics}, 32(4):41, 2013.

\bibitem{cao2014facewarehouse}
C.~Cao, Y.~Weng, S.~Zhou, Y.~Tong, and K.~Zhou.
\newblock Facewarehouse: A {3D} facial expression database for visual
  computing.
\newblock {\em IEEE Transactions on Visualization and Computer Graphics},
  20(3):413--425, 2014.

\bibitem{clarke2011automatic}
L.~Clarke, M.~Chen, and B.~Mora.
\newblock Automatic generation of {3D} caricatures based on artistic
  deformation styles.
\newblock {\em IEEE Transactions on Visualization and Computer Graphics},
  17(6):808--821, 2011.

\bibitem{diebel2006representing}
J.~Diebel.
\newblock Representing attitude: Euler angles, unit quaternions, and rotation
  vectors.
\newblock {\em Matrix}, 58(15-16):1--35, 2006.

\bibitem{gao2016efficient}
L.~Gao, Y.-K. Lai, D.~Liang, S.-Y. Chen, and S.~Xia.
\newblock Efficient and flexible deformation representation for data-driven
  surface modeling.
\newblock {\em ACM Transactions on Graphics}, 35(5):158, 2016.

\bibitem{gao2017sparse}
L.~Gao, Y.-K. Lai, J.~Yang, L.-X. Zhang, L.~Kobbelt, and S.~Xia.
\newblock Sparse data driven mesh deformation.
\newblock {\em arXiv preprint arXiv:1709.01250}, 2017.

\bibitem{cvpr99}
D.~B. Graham and N.~M. Allinson.
\newblock Norm\({}^{\mbox{2}}\)-based face recognition.
\newblock In {\em IEEE Conference on Computer Vision and Pattern Recognition},
  pages 1586--1591, 1999.

\bibitem{eigenweb}
G.~Guennebaud, B.~Jacob, et~al.
\newblock Eigen v3.
\newblock http://eigen.tuxfamily.org, 2010.

\bibitem{guo2017photo}
Y.~Guo, J.~Zhang, J.~Cai, B.~Jiang, and J.~Zheng.
\newblock {CNN}-based real-time dense face reconstruction with inverse-rendered
  photo-realistic face images.
\newblock {\em arXiv preprint arXiv:1708.00980}, 2017.

\bibitem{han2017deepsketch2face}
X.~Han, C.~Gao, and Y.~Yu.
\newblock {DeepSketch2Face}: a deep learning based sketching system for {3D}
  face and caricature modeling.
\newblock {\em {ACM} Transactions on Graphics}, 36(4):126:1--126:12, 2017.

\bibitem{huo2017webcaricature}
J.~Huo, W.~Li, Y.~Shi, Y.~Gao, and H.~Yin.
\newblock {WebCaricature}: a benchmark for caricature face recognition.
\newblock {\em arXiv preprint arXiv:1703.03230}, 2017.

\bibitem{jackson2017large}
A.~S. Jackson, A.~Bulat, V.~Argyriou, and G.~Tzimiropoulos.
\newblock Large pose {3D} face reconstruction from a single image via direct
  volumetric {CNN} regression.
\newblock {\em International Conference on Computer Vision}, 2017.

\bibitem{jiang20173d}
L.~Jiang, J.~Zhang, B.~Deng, H.~Li, and L.~Liu.
\newblock {3D} face reconstruction with geometry details from a single image.
\newblock {\em arXiv preprint arXiv:1702.05619}, 2017.

\bibitem{kemelmacher20113d}
I.~Kemelmacher-Shlizerman and R.~Basri.
\newblock {3D} face reconstruction from a single image using a single reference
  face shape.
\newblock {\em IEEE Transactions on Pattern Analysis and Machine Intelligence},
  33(2):394--405, 2011.

\bibitem{kemelmacher20083d}
I.~Kemelmacher-Shlizerman, R.~Basri, and B.~Nadler.
\newblock {3D} shape reconstruction of {Mooney} faces.
\newblock In {\em IEEE Conference on Computer Vision and Pattern Recognition},
  2008.

\bibitem{king2009dlib}
D.~E. King.
\newblock Dlib-ml: A machine learning toolkit.
\newblock {\em Journal of Machine Learning Research}, 10(Jul):1755--1758, 2009.

\bibitem{liang2002example}
L.~Liang, H.~Chen, Y.~Xu, and H.~Shum.
\newblock Example-based caricature generation with exaggeration.
\newblock In {\em Pacific Conference on Computer Graphics and Applications},
  pages 386--393, 2002.

\bibitem{liao2004automatic}
P.-Y. C. W.-H. Liao and T.-Y. Li.
\newblock Automatic caricature generation by analyzing facial features.
\newblock In {\em Asian Conference on Computer Vision}, volume~2, 2004.

\bibitem{liu2009semi}
J.~Liu, Y.~Chen, C.~Miao, J.~Xie, C.~X. Ling, X.~Gao, and W.~Gao.
\newblock Semi-supervised learning in reconstructed manifold space for 3d
  caricature generation.
\newblock {\em Computer Graphics Forum}, 28(8):2104--2116, 2009.

\bibitem{Shum-CVPR03}
Z.~Liu, H.~Chen, and H.~Shum.
\newblock An efficient approach to learning inhomogeneous {Gibbs} model.
\newblock In {\em {IEEE} Computer Society Conference on Computer Vision and
  Pattern Recognition}, pages 425--431, 2003.

\bibitem{more1978levenberg}
J.~J. Mor{\'e}.
\newblock The {Levenberg-Marquardt} algorithm: implementation and theory.
\newblock In {\em Numerical Analysis}, pages 105--116. Springer, 1978.

\bibitem{o19993d}
A.~J. O'Toole, T.~Price, T.~Vetter, J.~C. Bartlett, and V.~Blanz.
\newblock {3D} shape and {2D} surface textures of human faces: The role of
  ``averages'' in attractiveness and age.
\newblock {\em Image and Vision Computing}, 18(1):9--19, 1999.

\bibitem{o1997three}
A.~J. O'toole, T.~Vetter, H.~Volz, and E.~M. Salter.
\newblock Three-dimensional caricatures of human heads: Distinctiveness and the
  perception of facial age.
\newblock {\em Perception}, 26(6):719--732, 1997.

\bibitem{paysan20093d}
P.~Paysan, R.~Knothe, B.~Amberg, S.~Romdhani, and T.~Vetter.
\newblock A {3D} face model for pose and illumination invariant face
  recognition.
\newblock In {\em IEEE International Conference on Advanced Video and Signal
  based Surveillance}, pages 296--301, 2009.

\bibitem{RHODES1987473}
G.~Rhodes, S.~Brennan, and S.~Carey.
\newblock Identification and ratings of caricatures: Implications for mental
  representations of faces.
\newblock {\em Cognitive Psychology}, 19(4):473 -- 497, 1987.

\bibitem{Sadimon10survey}
S.~B. Sadimon, M.~S. Sunar, D.~B. Mohamad, and H.~Haron.
\newblock Computer generated caricature: A survey.
\newblock In {\em International Conference on CyberWorlds}, pages 383--390,
  2010.

\bibitem{sela2015computational}
M.~Sela, Y.~Aflalo, and R.~Kimmel.
\newblock Computational caricaturization of surfaces.
\newblock {\em Computer Vision and Image Understanding}, 141:1--17, 2015.

\bibitem{CVPR91}
M.~A. Shackleton and W.~J. Welsh.
\newblock Classification of facial features for recognition.
\newblock In {\em {IEEE} Conference on Computer Vision and Pattern
  Recognition}, pages 573--579, 1991.

\bibitem{sorkine2005laplacian}
O.~Sorkine.
\newblock Laplacian mesh processing.
\newblock In {\em Eurographics (STARs)}, pages 53--70, 2005.

\bibitem{sorkine2007rigid}
O.~Sorkine and M.~Alexa.
\newblock As-rigid-as-possible surface modeling.
\newblock In {\em Eurographics Symposium on Geometry Processing}, pages
  109--116, 2007.

\bibitem{sumner2004deformation}
R.~W. Sumner and J.~Popovi{\'c}.
\newblock Deformation transfer for triangle meshes.
\newblock {\em ACM Transactions on Graphics}, 23(3):399--405, 2004.

\bibitem{vlasic2005face}
D.~Vlasic, M.~Brand, H.~Pfister, and J.~Popovi{\'c}.
\newblock Face transfer with multilinear models.
\newblock {\em ACM Transactions on Graphics}, 24(3):426--433, 2005.

\bibitem{CGF:CGF1804}
S.~Wang and S.~Lai.
\newblock Manifold-based {3D} face caricature generation with individualized
  facial feature extraction.
\newblock {\em Computer Graphics Forum}, 29(7):2161--2168, 2010.

\bibitem{yu2004mesh}
Y.~Yu, K.~Zhou, D.~Xu, X.~Shi, H.~Bao, B.~Guo, and H.-Y. Shum.
\newblock Mesh editing with {Poisson}-based gradient field manipulation.
\newblock {\em ACM Transactions on Graphics}, 23(3):644--651, 2004.

\bibitem{zhu2015high}
X.~Zhu, Z.~Lei, J.~Yan, D.~Yi, and S.~Z. Li.
\newblock High-fidelity pose and expression normalization for face recognition
  in the wild.
\newblock In {\em IEEE Conference on Computer Vision and Pattern Recognition},
  pages 787--796, 2015.

\end{thebibliography}
}

\end{document}